\documentclass[10pt,twocolumn,letterpaper]{article}

\usepackage{cvpr}
\usepackage{times}
\usepackage{epsfig}
\usepackage{graphicx}
\usepackage{amsmath}
\usepackage{amssymb}
\usepackage{epstopdf}
\usepackage{color}
\usepackage{dblfloatfix} 

\usepackage[pagebackref=true,breaklinks=true,letterpaper=true,colorlinks,bookmarks=false]{hyperref}

\cvprfinalcopy % *** Uncomment this line for the final submission

\def\cvprPaperID{9} % *** Enter the CVPR Paper ID here

\ifcvprfinal\fi

\begin{document}

%%%%%%%%% TITLE
\title{Rank Persistence: Assessing the Temporal Performance \\of Real-World Person Re-Identification}

\author{Srikrishna Karanam, Eric Lam, and Richard J. Radke\\
Department of Electrical, Computer, and Systems Engineering\\
Rensselaer Polytechnic Institute\\
Troy, NY, United States\\
{\tt\small srikrishna@ieee.org, rjradke@ecse.rpi.edu}
}

\maketitle
%\thispagestyle{empty}

%%%%%%%%% ABSTRACT
\begin{abstract}
Designing useful person re-identification systems for real-world applications requires attention to operational aspects not typically considered in academic research.  Here, we focus on the temporal aspect of re-identification; that is, instead of finding a match to a probe person of interest in a fixed candidate gallery, we consider the more realistic scenario in which the gallery is continuously populated by new candidates over a long time period.  A key question of interest for an operator of such a system is: how long is a correct match to a probe likely to remain in a rank-k shortlist of possible candidates?  We propose to distill this information into a Rank Persistence Curve (RPC), which allows different algorithms' temporal performance characteristics to be directly compared.  We present examples to illustrate the RPC using  a new long-term dataset with multiple candidate reappearances, and discuss considerations for future re-identification research that explicitly involves temporal aspects.
\end{abstract}

%%%%%%%%% BODY TEXT
\section{Introduction}

Research in the area of automatic human re-identification, or re-id, has exploded in the past ten years.  The re-id problem is generally stated as: given an image of a person of interest as seen in a ``probe" camera view, how can we find the same person among a set of candidate people seen in a ``gallery" camera view?   Re-id research to date typically falls into one or more of the following categories:

\begin{itemize}
\item Appearance modeling, in which the goal is to design or learn a feature representation for re-id candidates that is invariant to challenges like viewpoint and illumination variation (e.g., \cite{bazzani2012symmetry, ma2012local, mclaughlin2016recurrent, satta2013appearance})
\item Metric learning, in which the goal is to learn, in a supervised fashion, a distance metric that is used to search for the person of interest in the gallery set (e.g., \cite{weinberger2009distance, prosser2010person, xiong2014person, liao2015person}).
\item Multi-shot re-id, in which both the probe and gallery candidates are represented as short image sequences instead of single frames (e.g., \cite{lisanti2015person,karanam2016person,li2015multi,wang2016person}).
\end{itemize}

\begin{figure*}[htbp!]
\centering
\includegraphics[scale=0.45]{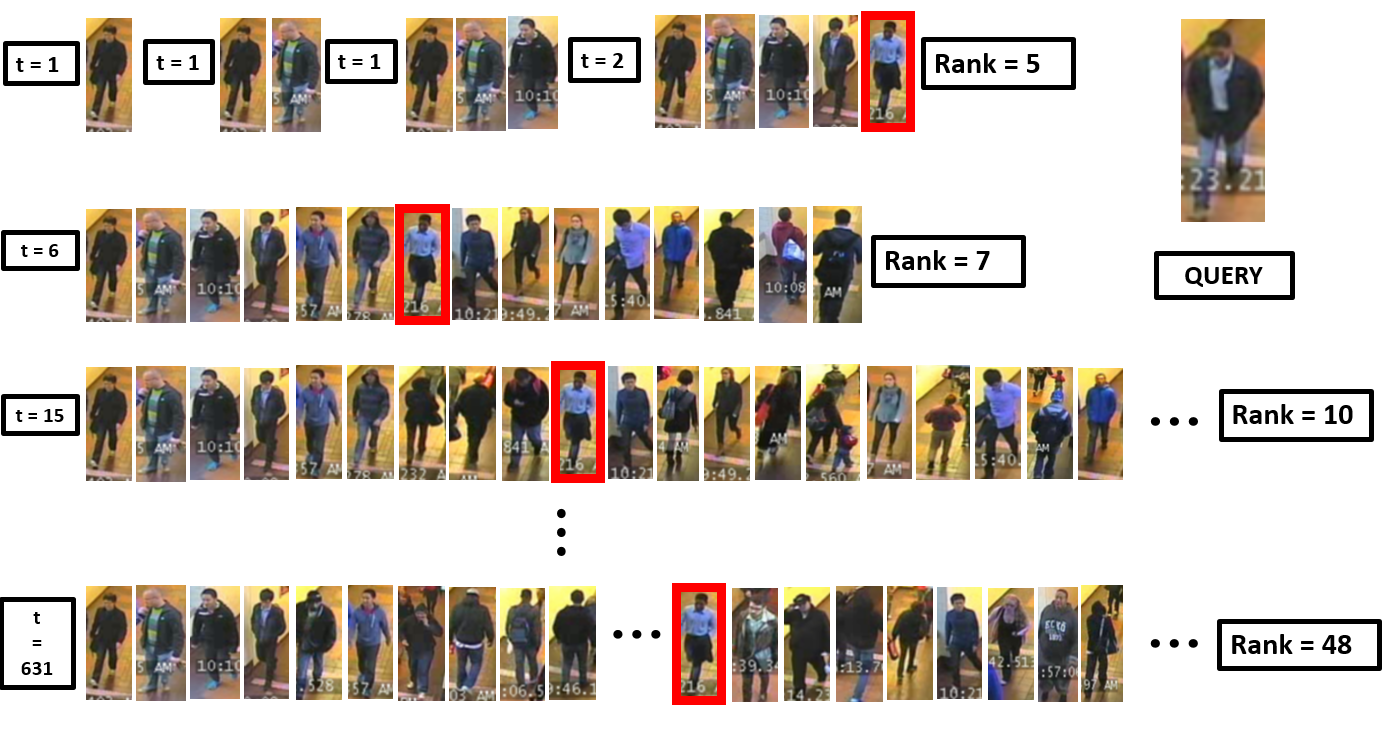}
\caption{An example showing the temporal evolution of a gallery set captured in a real surveillance system, and the corresponding time-varying rank for a given query
achieved by a re-id algorithm.  The candidate arrives at $t=2$ at rank 5, and drops in rank over time as more candidates arrive in the target camera.}
\label{fig:temporalGallery}
\end{figure*}

While critical to the success of a real-world deployment of a re-identification algorithm, research in these areas generally over-simplifies the problem that would face a real-world user of a re-id system.  In particular, the \emph{temporal} aspect of the re-id problem is totally ignored in most academic re-id research.  That is, in the real world, candidates  would be constantly added to the gallery as new subjects are automatically tracked, as opposed to presented to an algorithm all at once.  Even if a correct match to the probe appears in a rank-ordered shortlist shortly after they appear in a gallery camera, this isn't helpful to a user if the candidate is immediately ``pushed off'' the list after a few minutes by a new wave of incoming candidates.  To the user, a natural question is, how long can a correct match be expected to stay in the shortlist under typical circumstances?  Proposed re-id algorithms should be judged based on this notion of \emph{persistence} in time, not just raw batch performance as presented in a cumulative match characteristic (CMC) curve.  Figure \ref{fig:temporalGallery} illustrates the problem; an incoming correct match arrives at a low rank upon its first appearance, but is quickly pushed down the list as more candidates arrive.  In this example, the correct candidate would have only stayed on a shortlist of the top 10 candidates for 10 minutes (about the same duration as a typical coffee break!). 

In this paper, we explore several temporal aspects of re-id and propose new evaluation methodologies that allow different re-id algorithms to be compared based on the concept of rank persistence.  We discuss strategies for evaluating algorithms in circumstances when the same person of interest can appear multiple times in the gallery (which could occur if the gallery spans several hours or days) as well as when the performance on multiple persons of interest should be aggregated.  The key concept we propose is called the \emph{Rank Persistence Curve (RPC)}, which quantifies the percentage of candidates that remain at a certain rank for a given duration.  RPCs for different algorithms can be directly compared on the same dataset to allow a user to make informed choices about expected performance in real-world deployments.

We illustrate our proposed methodology with examples and experiments drawn from a dataset we collected over many hours at a light rail station, which contains multiple time-stamped reappearances of multiple actors.   We conclude the paper with further discussion about the temporal implications of re-id, and suggestions for future research in this area.

\section{Preliminaries}
\label{sec:dataAlgo}

Although the focus of this paper is not the proposal of a new re-id dataset or algorithm, we need instances of each to illustrate our proposed methodology.  In particular, since existing re-id benchmarking datasets lack temporal annotation, we collected a new dataset more suitable for this study.  

The dataset is composed of 10 hours of surveillance video data from a wall-mounted camera located in an indoor light rail station in the United States. We had 7 known actors participate in the data collection activity. Each actor appeared once in a morning session, and subsequently re-appeared in the same camera view in multiple afternoon and evening sessions. In each re-appearance, the actor wore slightly different clothes or accessories than in the previous session, to make the re-id problem more realistic and challenging.  We used each actor's appearance in the morning session as the ``probe" for subsequent re-id queries. In total, we ground-truthed 3 known re-appearances of each actor, each randomly spread across the 10-hour duration. 

\begin{figure*}[!htbp]
\centering
\includegraphics[scale=0.6]{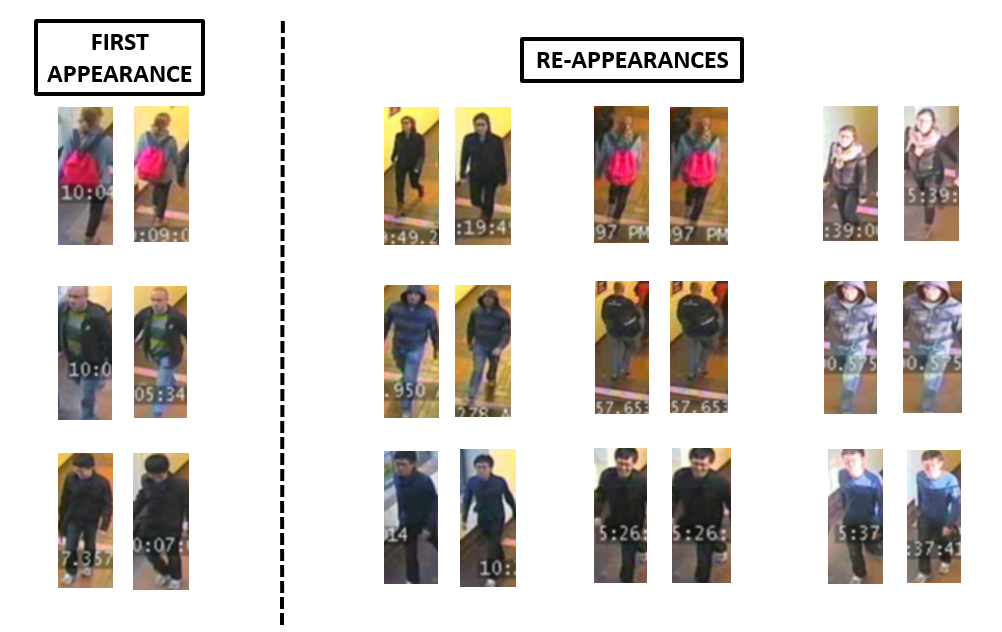}
\caption{Examples of actors and their appearances in our dataset.}
\label{fig:dataset}
\end{figure*}

To generate candidate tracks for the gallery (i.e., for both the actors and the distractor pedestrians that entered the camera view over the 10-hour video), we used an off-the-shelf person detector, based on the aggregated channel features (ACF) algorithm of Doll\'ar \etal \cite{dollar2010fastest, dollar2014fast}, to crop out person images. Sample images from the different appearances of the actors in our dataset are shown in Figure \ref{fig:dataset}. A statistical summary of our dataset is provided in Table \ref{tab:datasetSummary}. 

\begin{table}[!htb]
\centering
\caption{Statistical summary of our dataset}
\label{tab:datasetSummary}
\begin{tabular}{|c|c|}
\hline
Property & Total/Type  \\ \hline
\# Bounding Boxes & 4,639  \\
\# Known Actors & 7  \\
\# Known Actor Re-Appearances & 3 \\
\# Candidates (incl. actors) & 535 \\
\# Video Hours & 10  \\ 
Detection algorithm & ACF \cite{dollar2010fastest, dollar2014fast} \\ \hline
\end{tabular}%
\end{table}

As a baseline re-id algorithm, we used the recently proposed Gaussian of Gaussian (GOG) descriptor \cite{matsukawa2016hierarchical} to rank gallery candidates. GOG is an unsupervised algorithm that constructs an appearance model given image data and uses Euclidean distance to rank candidates.  The baseline algorithm is multi-shot in that it uses multiple frames of each candidate's appearance to form feature vectors to be compared, instead of a single rectangle.

Specifically, given a track of images for the probe appearance of the person of interest and each gallery candidate, we extract features for each image using GOG. Let $\mathbf{x}_{p}^{k}$, $k=1,\cdots, n$  and $\mathbf{x}_i^{k}$, $k=1, \cdots, m$ denote the $n$ feature vectors of the probe and the $m$ feature vectors of the $i^{th}$ gallery candidate, respectively. We then determine the appearance model for the probe, $\mathbf{f}_{p}$, and the gallery candidate, $\mathbf{f}_{i}$, as the mean feature vector of the available feature vectors. Specifically,

\begin{equation*}
\mathbf{f}_{p} = \frac{1}{n}\sum_{k=1}^{n}\mathbf{x}_{p}^{k}
\end{equation*}

\begin{equation*}
\mathbf{f}_{i} = \frac{1}{m}\sum_{k=1}^{m}\mathbf{x}_i^{k}
\end{equation*}

The similarity score $s_i$, used to rank gallery candidates, is computed using the Euclidean distance as

\begin{equation*}
s_{i}=\|\mathbf{f}_{p}-\mathbf{f}_{i}\| 
\end{equation*}

\section{Rank Persistence}

In this section, we present the concept of rank persistence in steps, working up to evaluating situations in which multiple persons of interest each appear multiple times over the course of a long video sequence.

\subsection{One probe, one reappearance}
\begin{figure}[htbp!]
\centering
\includegraphics[scale=0.3]{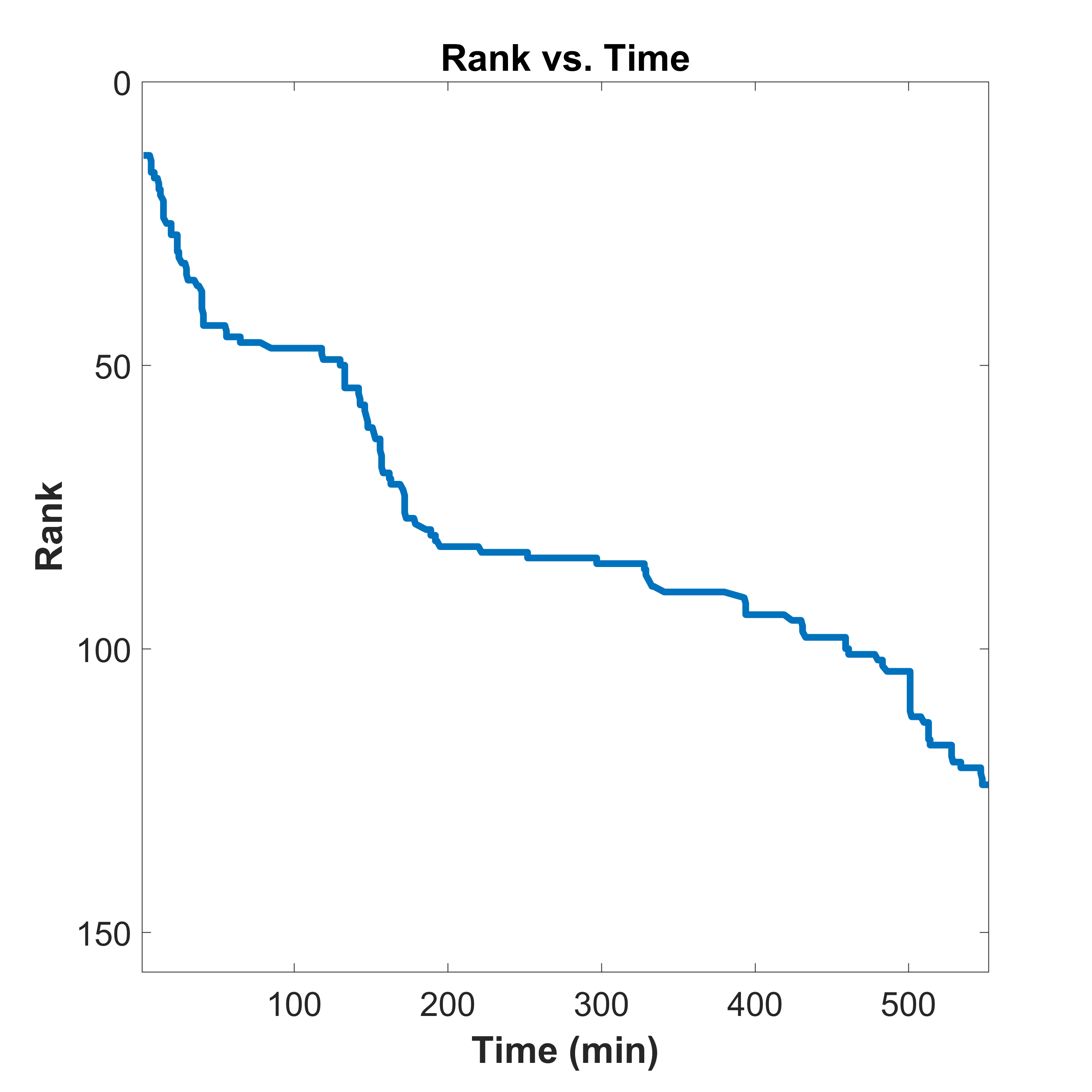}
\caption{An example of how the rank $r$ of a person of interest changes in a temporally-varying gallery set.  This candidate's first re-appearance is at $t=2$ minutes.}
\label{fig:samplePlot}
\end{figure}

We first consider the case of a single probe/query that has exactly one reappearance over the course of an entire video. In this case, we can easily graph their rank over time with respect to an ever-increasing gallery, as illustrated in Figure \ref{fig:samplePlot}.

The horizontal axis is real time (i.e., minutes from the beginning of our 10-hour test sequence).  The vertical axis shows the rank of the true reappearance in the gallery over time.  Clearly, the rank can only decrease as more candidates arrive in the gallery.  Also, the graph does not start at $t=0$ but indicates the time at which the reappearance occurs.  These per-probe curves will form the basis for the aggregate Rank Persistence Curve discussed in Section \ref{sec:rpc}.

\begin{figure*}[hbtp!]
\centering
\includegraphics[scale=0.45]{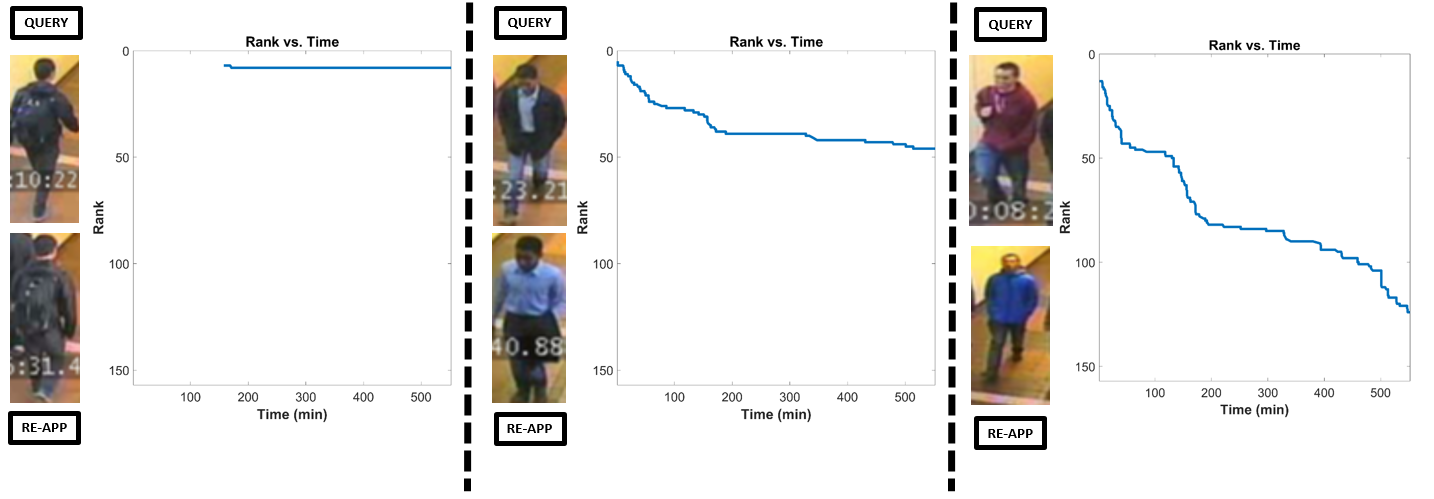}
\caption{Temporal rank curves for three sample queries.  In each case, the image on the top is the probe/query and the image on the bottom is the known reappearance in the gallery. The graph to the right of these images shows the temporal evolution of the rank of the known reappearance.}
\label{fig:qualitativeResults}
\end{figure*}

Figure \ref{fig:qualitativeResults} illustrates three rank curves for probe/gallery pairs of increasing difficulty.  In the first example, the gallery reappearance is very similar to the probe, and also dissimilar to other candidates seen over the course of the video, so the rank is fairly constant at a low value.  In the last example, the gallery reappearance is quite unlike the probe, so the rank increases quickly as new candidates appear.

\subsection{One probe, multiple reappearances}
\label{sec:multiplereap}

\begin{figure*}[!htbp]
\begin{center}
\includegraphics[scale=0.5]{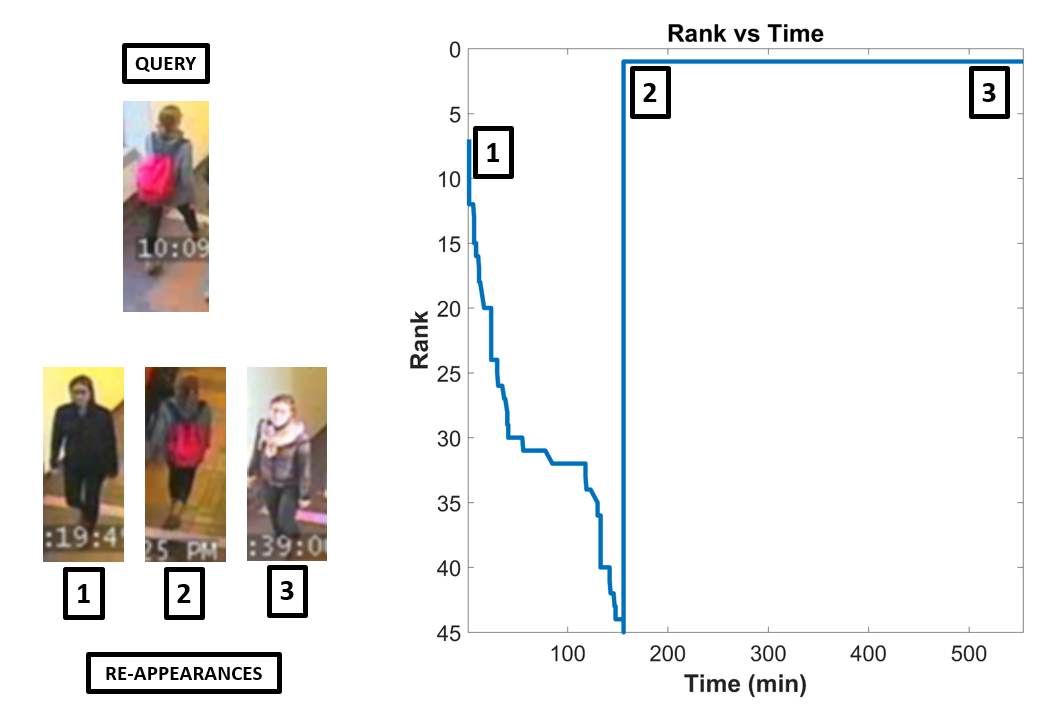}
\caption{Temporal rank plot for a query with multiple re-appearances.}
\label{fig:multiplereap}
\end{center}
\end{figure*}

In a real-world scenario, when we are searching for a  person of interest in streaming video over many hours or even days, it is possible that s/he may re-appear multiple times at different ranks.  Therefore, we modify the temporal rank curve from the previous section as illustrated in Figure ~\ref{fig:multiplereap}.  That is, the vertical axis shows the highest instantaneous rank held by \emph{any} reappearance of the probe.  

In this example, the first reappearance of the candidate in the gallery looks rather unlike the probe, so the initial rank is relatively high and increases sharply with time.  However, in the next reappearance, the gallery image looks very much like the probe; the rank of this new candidate is low, and the candidate persists at this rank for the rest of the long video.  A third reappearance of the candidate occurs later in the video, but is not as similar as the second reappearance, so does not affect the temporal rank plot.

\subsection{Multiple probes, multiple reappearances} \label{sec:rpc}

Finally, we arrive at the most general situation, in which we characterize the performance of a given re-id algorithm across multiple probes, each of which may have multiple reappearances in the gallery.  We define the \emph{Rank Persistence Curve (RPC)} to evaluate performance as follows.  First, we fix a specific rank $r$.  For each duration $d$ (in real units) on the horizontal axis, we plot the percentage of candidates that appear continuously in the top-$r$ list for at least $d$ units. Thus, the RPC is monotonically decreasing, and RPCs at higher ranks dominate those at lower ranks.  

\begin{figure*}[!htbp]
\centering
\begin{tabular}{cc}
\includegraphics[scale=0.37]{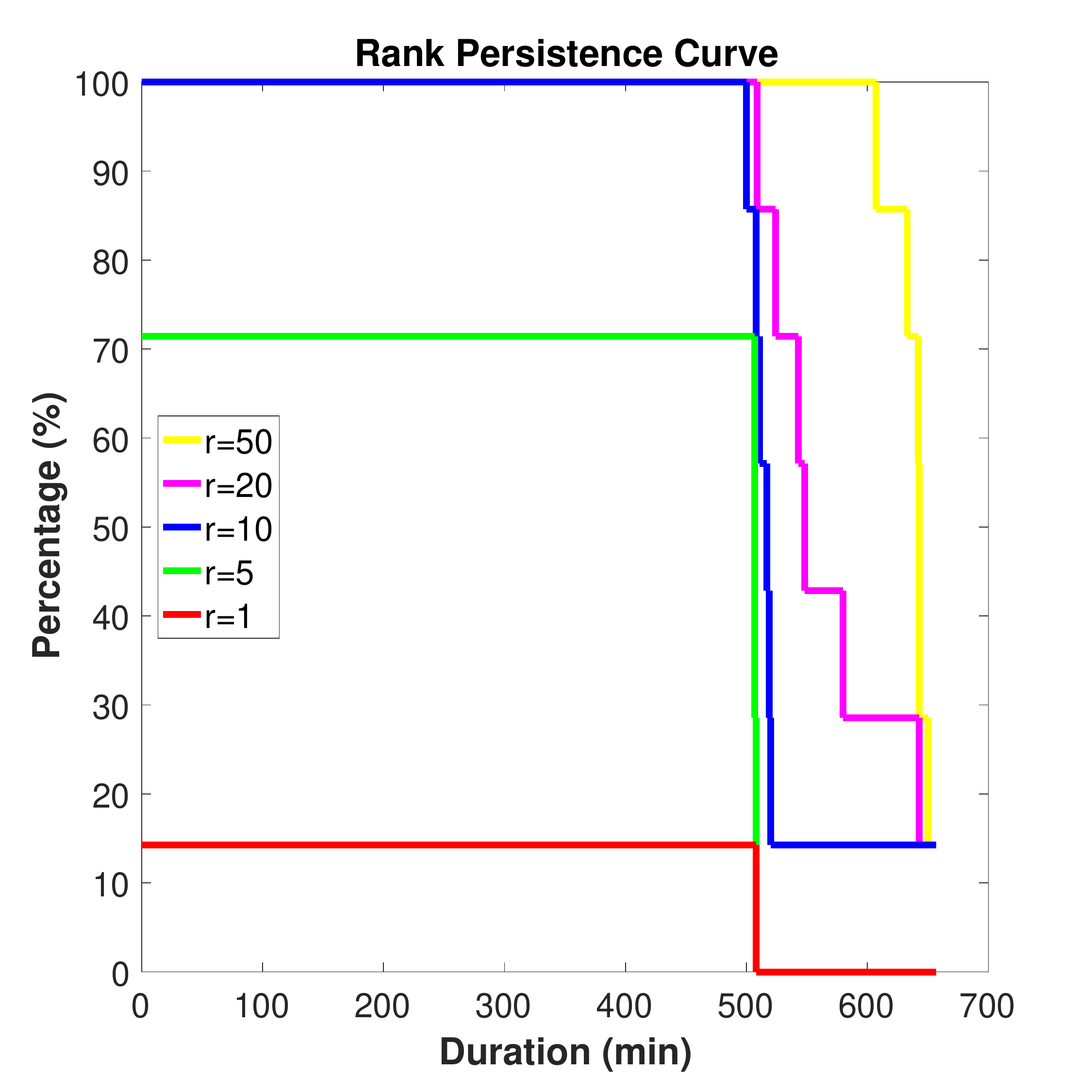} & 
\includegraphics[scale=0.37]{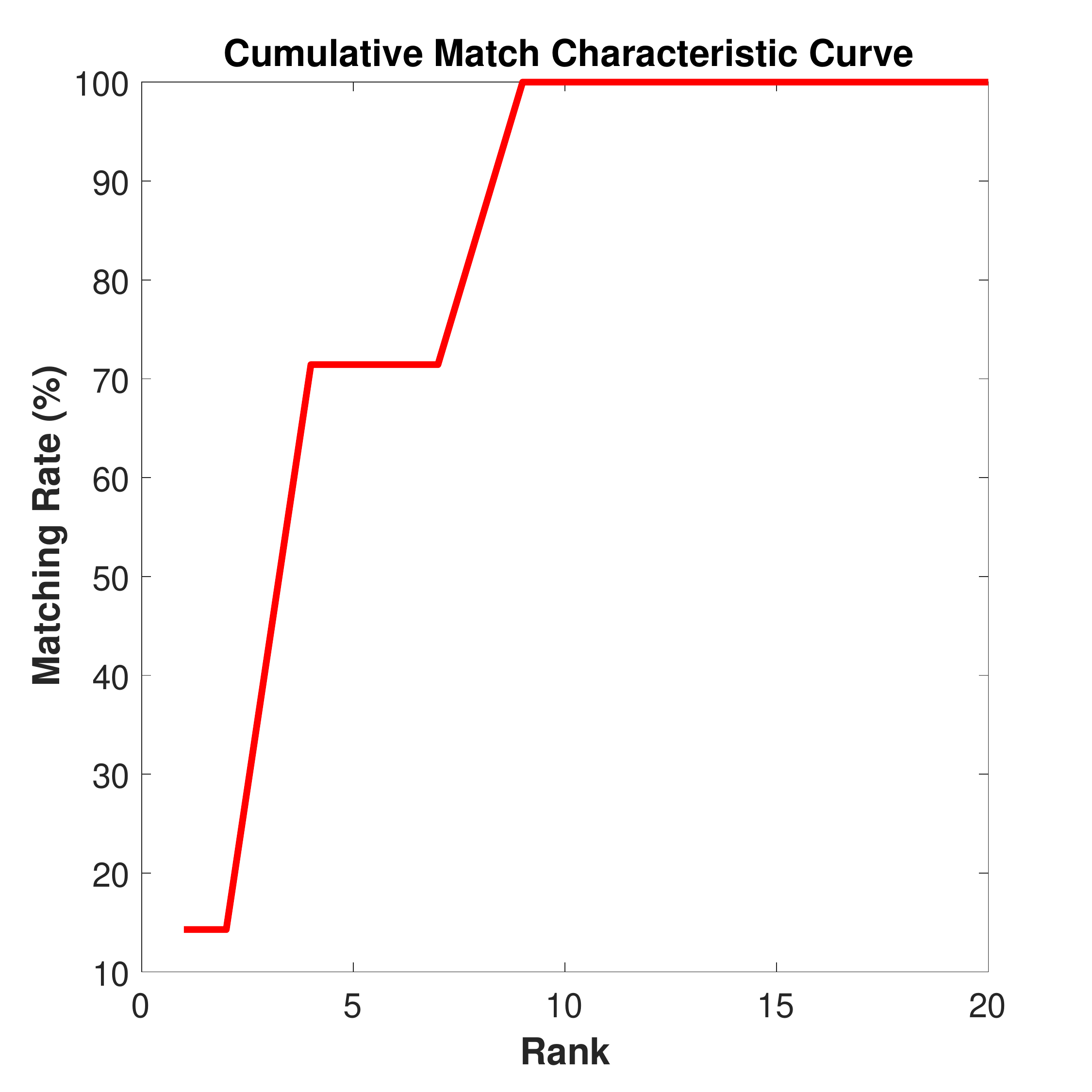} \\
(a) & (b)
\end{tabular}
\caption{(a) The proposed Rank Persistence Curve (RPC) at different ranks for our dataset/baseline re-id algorithm combination.  (b) The traditional Cumulative Match Characteristic (CMC) curve for our dataset/baseline re-id algorithm combination.}
\label{fig:sampleRPCCMC}
\end{figure*}

Figure \ref{fig:sampleRPCCMC}a illustrates RPCs for our dataset/baseline re-id algorithm combination.  In contrast, a traditional Cumulative Match Characteristic (CMC) curve, also for our dataset/baseline re-id algorithm combination (averaging all available feature vectors for every candidate), is shown in Figure \ref{fig:sampleRPCCMC}b. We can see that the two types of curves are qualitatively different. Since we want to capture the temporal aspect of rank in the RPC, the dependent axis is no longer rank but duration, and we need a third ``axis'' (in this case color) to indicate rank. 

Let's focus on the RPC for $r=1$, shown in red.  This captures our objective of visualizing how likely and how long a candidate is to stay at rank 1 across a long video sequence.  Clearly, this is a stringent requirement, and we can see that only 1 of the 7 probes ever had a reappearance at rank 1 at all.  If we consider the RPC for $r=5$, we see that more candidates are likely to persist at rank 5 for longer amounts of time.  RPCs can help operators of re-id systems answer questions like, can my re-id algorithm  be expected to preserve correct matches in the top-10 shortlist for at least 15 minutes at least 90\% of the time?  These considerations are important both in terms of the length of the shortlist (real-world end users, typically not computer vision experts, would not want to scroll through pages and pages of candidates to find the person of interest) and the duration of persistence (in real-world scenarios, end users may only get around to checking the output of a re-id surveillance system a few times an hour).

\begin{figure}[!htbp]
\centering
\includegraphics[scale=0.35]{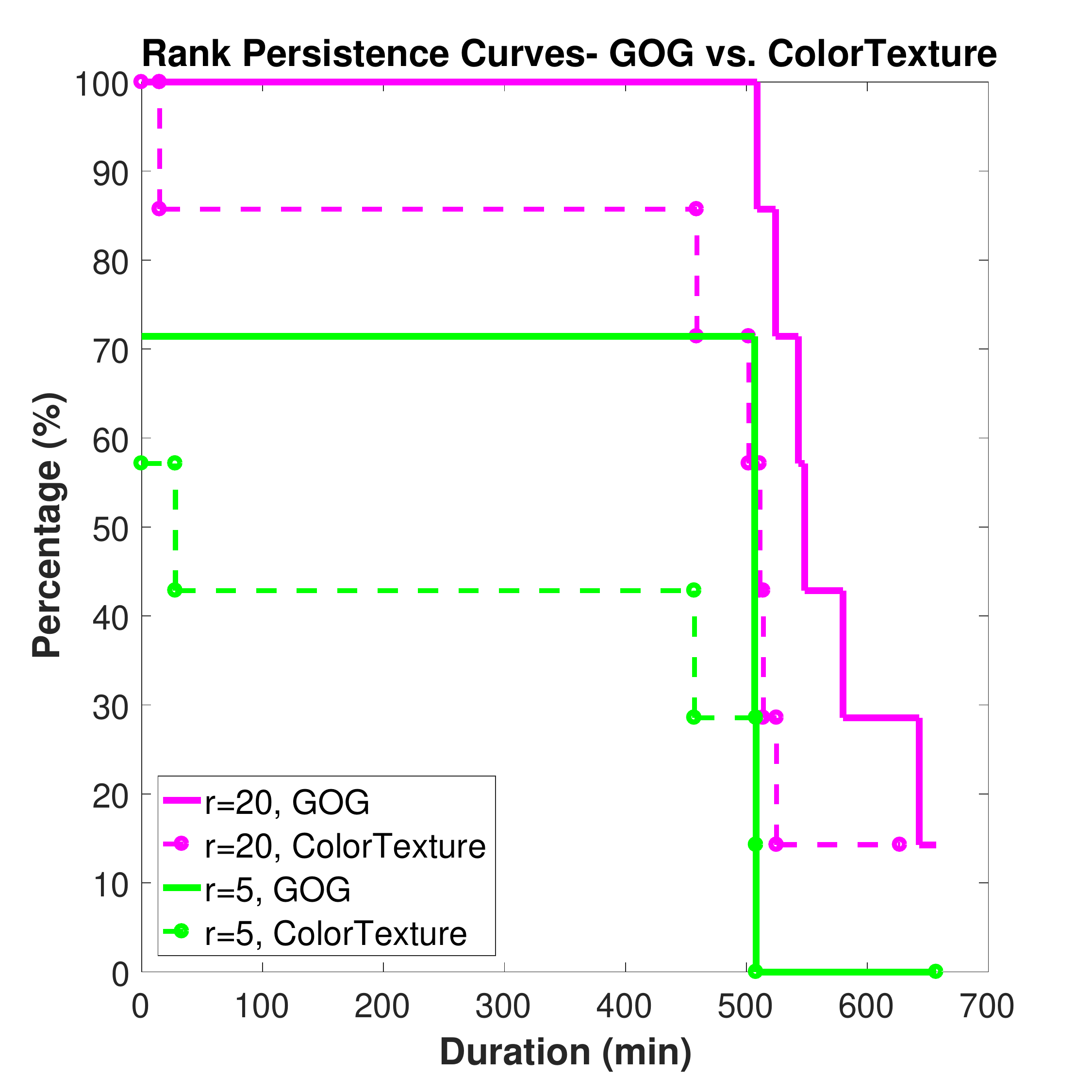}
\caption{RPCs can be used to easily compare the temporal performance of competing re-id algorithms. In this example, the GOG algorithm's RPC (solid curve) dominates the ColorTexture's algorithm's RPC (dashed curve) at $r=5$ and $r=20$, demonstrating the superiority of GOG from a temporal persistence perspective.}
\label{fig:RPC_gog_colortex}
\end{figure}

Thus, RPCs can be used to quickly visualize the performance of competing re-id systems applied in the same environment.  For example, Figure \ref{fig:RPC_gog_colortex} shows the RPCs for our baseline algorithm using GOG vs.~the ColorTexture algorithm proposed by Gray and Tao \cite{gray2008viewpoint}, which uses an ensemble of localized color and texture features. We can see that the GOG algorithm dominates the older algorithm with respect to temporal persistence, and that the RPCs capture the essence of both algorithms' performance in an easy-to-read graph.

\subsection{Video Flow Density}
\label{sec:density}
We conclude this section with a brief discussion on video flow density, a topic we have not considered in this paper but has a direct impact on the behavior of the rank of the person of interest. We define video flow density as the instantaneous number of people seen by the gallery camera per unit time. To understand how this might impact rank persistence, consider two nonoverlapping time blocks with the same duration. In the first block, the gallery camera sees $x$ people walking by. In the second block, the gallery camera sees $y$ people walking by.  If $x \gg y$, we add many more people in the gallery in the first time block, likely resulting in a steeper increase in a candidate's re-id rank compared to the second time block. This suggests we should pay close attention to the rate at which new people in the video are added to the gallery. 

\section{Discussion and Future Work}

Now that the underlying computer vision and machine learning technologies for re-id have matured \cite{karanam2016systematic,zheng2016person}, we contend that researchers should begin to take a broader view of evaluating how re-id algorithms should integrate into functional real-world systems.  For example, as noted in Camps et al.~\cite{camps2016from}, the problem of comparing one candidate rectangle of pixels to another is only a small part of a fully automated re-id system.  Instead, we must take into account that the candidate rectangles are likely generated by an automatic (and possibly inaccurate) human detection and tracking subsystem, that the overall system needs to operate in real time, and that the system may be in operation for very long periods of time.  Instead of benchmarking datasets in which the gallery images are acquired only a few moments after the probe images, we must consider crime prevention applications in which a perpetrator may return to the scene of the crime days after their initial detection.  In such cases, the gallery of candidates is ever-expanding, and for long periods of time may not contain the person of interest at all.

These real-world considerations present several challenges to the re-id research community, several of which we discuss below.  

\begin{itemize}
\item While they are derived from real-world surveillance video, current re-id benchmarking datasets such as VIPeR \cite{gray2008viewpoint}, iLIDS-VID \cite{wang2014person}, and MARS \cite{zheng2016mars} lack time stamps for the gallery sets, rendering them inappropriate for the type of temporal research discussed here.  While it might be possible to repurpose them for temporal re-id research, adding artificial time stamps seems suboptimal.  It would be better to generate a new temporal re-id dataset for this purpose, ideally spanning several days and thousands of tracked candidates.  For validation, multiple reappearances of a substantial number of actors would need to be included.  The dataset discussed in this paper is a step in this direction, but has too few unique actors to be broadly useful.

\item The concept of ``splits'' is critical for fair comparisons of re-id algorithms using current benchmarking datasets.  That is, the provider of the dataset typically specifies a random subset of the data to be used for training and another to be used for testing.  A similar concept would need to be developed for temporal re-id datasets, i.e., not only specifying which candidates should be used for training/testing, but also which temporal spans should be used in the time-evolving galleries.

\item As discussed in Section \ref{sec:density}, the x-axis of the proposed Rank Persistence Curves is critically coupled to the video flow density of candidates in the underlying camera.  Operationally, it would be more useful to characterize rank persistence in real units of minutes or hours; however, this ties the RPC strongly to a particular dataset and possibly a time of day.  For example, different rail stations may have different levels of traffic, and the traffic would vary during rush hour vs.~off-peak times.  Temporal re-id benchmarking protocols would need to distinguish subsets of data with respect to video flow rate.

\item In our experience with integrating academic re-id algorithms into operational surveillance command centers, we found the issue of user interfaces to be extremely important.  The similarity between a rank-$k$ shortlist and a police lineup was an effective analogy.  However, the potentially very long time scales for crime prevention applications requires the re-evaluation of an operationally meaningful shortlist.  Should candidates ``age out'' of the ranked list using some sort of forgetting factor?  Should extremely promising candidates from long ago be kept alongside less-certain but more timely recent candidates?  Should the time-varying gallery contain all candidates ever seen or only those from the last $N$ minutes?  These considerations require close consultation with the potential users of the system to understand and set expectations and corresponding interface choices.

\end{itemize}

\section*{Acknowledgements}
This material is based upon work supported by the U.S. Department of Homeland Security, Science and Technology Directorate, Office of University Programs, under Grant Award 2013-ST-061-ED0001. The views and conclusions contained in this document are those of the authors and should not be interpreted as necessarily representing the official policies, either expressed or implied, of the U.S. Department of Homeland Security. Thanks to John Joyce for supplying the video data.

{\small
\bibliographystyle{ieee}
\bibliography{cvpr2017_bib}

\begin{thebibliography}{10}\itemsep=-1pt

\bibitem{bazzani2012symmetry}
L.~Bazzani, M.~Cristani, and V.~Murino.
\newblock Symmetry-driven accumulation of local features for human
  characterization and re-identification.
\newblock {\em Comput. Vision and Image Understanding (CVIU)}, 117(2):130--144,
  2013.

\bibitem{camps2016from}
O.~Camps, M.~Gou, T.~Hebble, S.~Karanam, O.~Lehmann, Y.~Li, R.~Radke, Z.~Wu,
  and F.~Xiong.
\newblock From the lab to the real world: Re-identification in an airport
  camera network.
\newblock {\em IEEE Transactions on Circuits and Systems for Video Technology
  (T-CSVT)}, 27(3):540--553, 2017.

\bibitem{dollar2014fast}
P.~Doll{\'a}r, R.~Appel, S.~Belongie, and P.~Perona.
\newblock Fast feature pyramids for object detection.
\newblock {\em IEEE Trans. Pattern Anal. Mach. Intell. (T-PAMI)},
  36(8):1532--1545, 2014.

\bibitem{dollar2010fastest}
P.~Doll{\'a}r, S.~J. Belongie, and P.~Perona.
\newblock The fastest pedestrian detector in the west.
\newblock In {\em Proc. Brit. Mach. Vision Conf. (BMVC)}, 2010.

\bibitem{gray2008viewpoint}
D.~Gray and H.~Tao.
\newblock Viewpoint invariant pedestrian recognition with an ensemble of
  localized features.
\newblock In {\em Eur. Conf. Comput. Vision (ECCV)}. 2008.

\bibitem{karanam2016systematic}
S.~Karanam, M.~Gou, Z.~Wu, A.~Rates{-}Borras, O.~I. Camps, and R.~J. Radke.
\newblock A comprehensive evaluation and benchmark for person
  re-identification: Features, metrics, and datasets.
\newblock {\em arXiv preprint arXiv:1605.09653}, 2016.

\bibitem{karanam2016person}
S.~Karanam, Y.~Li, and R.~J. Radke.
\newblock Person re-identification with block sparse recovery.
\newblock {\em Image and Vision Computing}, 60:75--90, 2017.

\bibitem{li2015multi}
Y.~Li, Z.~Wu, S.~Karanam, and R.~J. Radke.
\newblock Multi-shot human re-identification using adaptive fisher discriminant
  analysis.
\newblock In {\em Proc. Brit. Mach. Vision Conf. (BMVC)}, 2015.

\bibitem{liao2015person}
S.~Liao, Y.~Hu, X.~Zhu, and S.~Z. Li.
\newblock Person re-identification by local maximal occurrence representation
  and metric learning.
\newblock In {\em IEEE Conf. Comput. Vision and Pattern Recognition (CVPR)},
  2015.

\bibitem{lisanti2015person}
G.~Lisanti, I.~Masi, A.~D. Bagdanov, and A.~Del~Bimbo.
\newblock Person re-identification by iterative re-weighted sparse ranking.
\newblock {\em IEEE Trans. Pattern Anal. Mach. Intell. (T-PAMI)},
  37(8):1629--1642, 2015.

\bibitem{ma2012local}
B.~Ma, Y.~Su, and F.~Jurie.
\newblock Local descriptors encoded by {F}isher vectors for person
  re-identification.
\newblock In {\em ECCV Workshops}, 2012.

\bibitem{matsukawa2016hierarchical}
T.~Matsukawa, T.~Okabe, E.~Suzuki, and Y.~Sato.
\newblock Hierarchical {G}aussian descriptor for person re-identification.
\newblock In {\em IEEE Conf. Comput. Vision and Pattern Recognition (CVPR)},
  2016.

\bibitem{mclaughlin2016recurrent}
N.~McLaughlin, J.~Martinez~del Rincon, and P.~Miller.
\newblock Recurrent convolutional network for video-based person
  re-identification.
\newblock In {\em IEEE Conf. Comput. Vision and Pattern Recognition (CVPR)},
  2016.

\bibitem{prosser2010person}
B.~Prosser, W.-S. Zheng, S.~Gong, and T.~Xiang.
\newblock Person re-identification by support vector ranking.
\newblock In {\em Proc. Brit. Mach. Vision Conf. (BMVC)}, 2010.

\bibitem{satta2013appearance}
R.~Satta.
\newblock Appearance descriptors for person re-identification: a comprehensive
  review.
\newblock {\em arXiv preprint arXiv:1307.5748}, 2013.

\bibitem{wang2014person}
T.~Wang, S.~Gong, X.~Zhu, and S.~Wang.
\newblock Person re-identification by video ranking.
\newblock In {\em Eur. Conf. Comput. Vision (ECCV)}. 2014.

\bibitem{wang2016person}
T.~Wang, S.~Gong, X.~Zhu, and S.~Wang.
\newblock Person re-identification by discriminative selection in video
  ranking.
\newblock {\em IEEE Trans. Pattern Anal. Mach. Intell. (T-PAMI)},
  38(12):2501--2514, 2016.

\bibitem{weinberger2009distance}
K.~Q. Weinberger and L.~K. Saul.
\newblock Distance metric learning for large margin nearest neighbor
  classification.
\newblock 10:207--244, 2009.

\bibitem{xiong2014person}
F.~Xiong, M.~Gou, O.~Camps, and M.~Sznaier.
\newblock Person re-identification using kernel-based metric learning methods.
\newblock In {\em Eur. Conf. Comput. Vision (ECCV)}. 2014.

\bibitem{zheng2016mars}
L.~Zheng, Z.~Bie, Y.~Sun, J.~Wang, C.~Su, S.~Wang, and Q.~Tian.
\newblock {MARS}: A video benchmark for large-scale person re-identification.
\newblock In {\em Eur. Conf. Comput. Vision (ECCV)}, 2016.

\bibitem{zheng2016person}
L.~Zheng, Y.~Yang, and A.~G. Hauptmann.
\newblock Person re-identification: Past, present and future.
\newblock {\em arXiv preprint arXiv:1610.02984}, 2016.

\end{thebibliography}
}

\end{document}